\documentclass{article}

\PassOptionsToPackage{numbers, compress}{natbib}

\usepackage[final]{nips_2018}
\usepackage{amsmath,amssymb,amsfonts}
\usepackage{bbm}
\usepackage{graphicx}
\usepackage[english]{babel}
\usepackage{blindtext}
\usepackage{tabularx}

\usepackage[caption=false, subrefformat=parens, labelformat=parens]{subfig}

\usepackage[utf8]{inputenc} 
\usepackage[T1]{fontenc}    
\usepackage{hyperref}       
\usepackage{url}            
\usepackage{booktabs}       
\usepackage{amsfonts}       
\usepackage{nicefrac}       
\usepackage{microtype}      

\usepackage{longtable}

\newcounter{daggerfootnote}
\newcommand*{\daggerfootnote}[1]{%
    \setcounter{daggerfootnote}{\value{footnote}}%
    \renewcommand*{\thefootnote}{\fnsymbol{footnote}}%
    \footnote[2]{#1}%
    \setcounter{footnote}{\value{daggerfootnote}}%
    \renewcommand*{\thefootnote}{\arabic{footnote}}%
    }

\title{Emulating malware authors for proactive protection using GANs over a distributed image visualization of dynamic file behavior}

\author{
	Vineeth S.~Bhaskara\thanks{Alternative Email: \href{mailto:bhaskaravineeth@gmail.com}{\texttt{bhaskaravineeth@gmail.com}}}\\
	Symantec Corporation\\
	Pune, India 411 014 \\
	\texttt{vineeth\_bhaskara@symantec.com} \And 
	 Debanjan Bhattacharyya\thanks{\href{mailto:b.debanjan@gmail.com}{\texttt{b.debanjan@gmail.com}}}\\
	 Symantec Corporation\\
	 Pune, India 411 014 \\
	 \texttt{debanjan\_bhattachary@symantec.com} \\
}

\begin{document}

\maketitle

\begin{abstract}
		Malware authors have always been at an advantage of being able to adversarially test and augment their malicious code, before deploying the payload, using  anti-malware products at their disposal. The anti-malware developers and threat experts, on the other hand, do not have  such a privilege of tuning  anti-malware products against zero-day attacks pro-actively. This allows the malware authors to being a step ahead of the anti-malware products, fundamentally biasing the cat and mouse game played by the two parties. In this paper, we propose a way that would enable machine learning based threat prevention models to bridge that gap by being able to tune against a deep generative adversarial network (GAN), which takes up the role of a malware author and generates new types of malware. The GAN is trained  over a reversible distributed  RGB image representation of known malware behaviors, encoding the sequence of API call ngrams and the corresponding term frequencies. The generated images represent synthetic malware that can be decoded back to the underlying API call sequence information. The image representation is not only demonstrated as a general technique of incorporating necessary priors for exploiting convolutional neural network architectures for generative or discriminative modeling, but also as a visualization method for easy manual software or malware categorization, by having individual API ngram information distributed across the image space. In addition, we also propose using smart-definitions for detecting malwares based on  perceptual hashing of these images. Such hashes are potentially more effective than cryptographic  hashes that do not carry any meaningful similarity metric, and hence, do not generalize well.
\end{abstract}

\section{Introduction}

With the growing influence of connectivity and machine intelligence in our day-to-day life, ranging from 
the use of smart personal assistants like Google Assistant, to peer-to-peer connected smart cars and IoT devices, the threat landscape has become more diverse  with increasing repercussions.  A successful breach can range from loosing of confidential data, loss of customers, to fatalities or even death. So, it is not  enough to be retroactive in protection. The existing anti-malware products mostly depend on highly specific malware-definitions based on cryptographic hashing algorithms. They use SHA256, as an example, for fingerprinting new malicious samples. However such samples reach the vendor's servers for analysis, only after some hours or days of it being first detected on a device.  Most of the harm might have already been done  by the time the anti-malware vendor gets the file for analysis. Instead, it is now important that the approach to stop malware must be increasingly proactive, where the anti-malware product constantly creates new malware behavior ``by itself'', intelligently, and discovers  unknown security holes in a system even before they get exploited by a breach. Such an automated way for creation of well-defined malware programs with unique behaviors, which would be as good as the programs created by humans, is  a difficult problem that not only involves modeling the logic, but also the semantics and the grammar of a programming language. We demonstrate, however, that if the malware behaviors can be approximately encoded into a representation, such as the image representation we propose, and given that, existing state-of-the-art generative models like GANs can train on them, then, one can sample   representations of new malware behavior from the trained generator network, emulating a malware author.

One of the key reasons for the success of deep learning technologies, in addition to the availability of better computational power at lower costs, has been the vast scope and research in architectural engineering that  constantly improved the training process and effective modeling of the priors of the problem. For example, for improving the training, the Residual Block architecture \cite{resnet} with its skip-like additive connections, ensures that each subsequent layer is at least as good as the previous, thereby, addressing  the gradient-vanishing problem in deeper networks. More tricks such as dropouts, non-saturating activation functions such as ReLU, first successfully demonstrated by Krizhevsky \textit{et al.} \cite{alexnet}, help in stabilizing the training process and improving generalization. An example of the architectural modifications to improve modeling the priors of a given problem, involves the idea of  weight-sharing, for instance, within the convolution operation \cite{convnet, convnet2} on image datasets. The weights per filter map are shared across an image to allow activation by similar localized features, and, therefore, provide translational invariance of the prediction against the semantic content of the input.
Similarly, weight-sharing across time, engineered in recurrent neural networks, enables better generalization for inputs involving sequential data of arbitrary lengths. Further, data-augmentation methods \cite{alexnet}, transformer networks \cite{transformer}, amongst others, are examples of architectural improvements for inputs with  invariance requirements involving more general Affine transformations, including, rotation.  Another important  work on data-driven architectural engineering includes modeling CNNs over datasets intrinsically structured as graphs. Since a graph is defined by the vertices and the local connections, such systems  require invariance of the prediction under graph isomorphism, in place of translation. For example, Duvenaud \textit{et al.} \cite{duvenaud} proposed CNNs over molecules with convolutions applied locally (locality defined by edges) after modeling each atom  as a node in a graph of arbitrary size and shape, predicting specific chemical properties of the molecules with a great accuracy. Advances in the architectures have also involved  intuition borrowed from other research fields. The recent capsule-based network architecture from Sabour \textit{et al.} \cite{capsnet}, for example, focuses on preserving the relative pose information of the visual entities in an image, by using units called ``capsules'', which encapsulate various relative spacial information of a feature in the form of a vector, as opposed to scalar outputs from pooling neurons, drawing references from psychology and computer graphics.

In contrast, there have not been such tailored end-to-end data-driven learning models on executables, understanding their behavior, till date. This could majorly be attributed to the difficulty of engineering the architectural modifications in a neural network, that would efficiently capture the priors of the executables appropriately. Unlike the case of images where the required invariances (like translation) are common to the entire data, for executables, the invariance of the bytes per say, varies with each file based on  programmer logic and the  optimizations that the compiler makes. For instance, a particular chunk of bytes in a file might represent a function definition that could be invoked from some other chunk of bytes. In such a case, it makes sense to require ``translational'' invariance of the overall behaviour (malicious or benign) with respect to those bytes defining the function. Since different files could have such byte chunks representing the same function definition, at different offsets, and have a non-linear ordering of instructions at run-time with the programmatic constructs such as loops, jumps, etc., it becomes very hard for existing deep learning models to generalize well from such data.

Some attempts towards this direction involve a couple of recent works from nVIDIA \cite{nvidia} and Microsoft Research \cite{msftcodeml}. The former work implements a 1D ConvNet  directly over the raw file binaries, which they refer to as MalConvNet, as a way of overcoming the memory decay in RNNs over long sequences, while the later one, models the programming constructs as graphs, building on the ideas of Duvenaud \textit{et al.} \cite{duvenaud}. But the efficacy of these models are far from what one expects to achieve when compared to the performance of deep learning models in other applications. Theoretically, predicting the  behavior of  files in arbitrary environments with static analysis of the raw binaries, is  limited, due to the Halting problem that is NP-hard for an optimal solution. Though this might not be restrictive for a practical (as opposed to optimal) end-to-end deep learning framework, it adds to the complexity of modeling over the raw bytes. 

Therefore, we propose taking a dynamic analysis approach in examining the file executables, that involves monitoring the program execution on a target environment as a sequence of the low-level API calls, which determines its overall high-level behavior. In contrast to static analysis, this approach is  resistant to any obfuscation of the payload by methods, such as, encryption of the corresponding raw bytes, which is  generally the case with advanced malwares.

In this paper, we propose a general method for visualizing a sequence of words (events or API calls) in the form of an image that is visually distinctive, where each pixel partly encodes the overall behavior of an execution sequence, losslessly. The image representation proposed incorporates the necessary priors required for exploiting the convolutional neural network architectures in the generative  (e.g. GANs) and the discriminative  (e.g. deep CNN-based malware classifiers) regimes. Moreover, it is also shown to be an effective visualization technique for easy manual software or malware categorization in Section \ref{sec:categorization}. Additionally, we also demonstrate how a deep Wasserstein GAN model with gradient penalty \cite{wgan2}, referred to as WGAN-GP, can be trained over the proposed image representation of the malicious file behaviors, to assume the role of a malware author and generate new malicious behavioral-sequences; interpolating, extrapolating or even combining the behavioral features of known malwares. Such new unseen malwares can then be used to test and enhance (behavioral) signature based anti-malware solutions proactively. 

An earlier work by Hu \textit{et al.} \cite{hu} highlights an application of GANs in evading detection by  generating adversarial inputs that fool existing machine learning based protection on clients. But for such an application to be practical, the malware authors must possess precise information on the features used by the  classifier, which is generally kept confidential by the security vendor. It is  hard for a malware author to figure out the features used, as the possible combinations (e.g. ngrams of the API calls for arbitrary $n$) to try out is huge. We, on the other hand, exploit GANs as  malware emulators for  training or testing malware classifiers  against the generated malwares for proactive protection. 

The rest of the paper is organized as sections in the following order:
\begin{itemize}
	\item  \hyperref[sec:imagegen]{Mapping file behaviors to RGB images}.
	\item  \hyperref[sec:categorization]{Applications on Software/Malware Categorization and Smart-Definitions}.
	\item \hyperref[sec:classifier]{Simple Malware Classifier trained on the behavior-mapped RGB images}.
	\item  \hyperref[sec:gan]{Modeling Malware Behaviors with Generative Adversarial Networks}.

\end{itemize}
Finally, we conclude by discussing on future directions, and  possible improvements over the initial implementation presented\daggerfootnote{References to the code, training parameters, dataset of files used, and the full resolution PNG figures are available at \url{https://github.com/bsvineethiitg/malwaregan}.}.

\section{Mapping file behaviors to RGB Images} \label{sec:imagegen}

The primary behavioral information obtained from an execution log, consists of the API (event or call) ngrams, their frequencies, and the sequence of their occurrences. For visualizing the executions, one would want to map this information to the pixel intensities of an image representation, that would allow easy manual categorization, and enable applying deep convolutional neural network-based architectures  across generative and discriminative models. A trivial way of mapping such information into an image domain, might involve encoding the individual ngram words with pixel positions, and the tf-idfs of those words as their intensities. Such a image representation is \textit{local} where a pixel corresponds to only a single feature of the behavioral trace, i.e., a single API call ngram.

We propose a \textit{distributed} image representation, rather than a local one, where each feature corresponding to a distinct ngram partly contributes to the intensity values across all the pixels of the image.

Some considerations that motivate a distributed image representation include:
\begin{enumerate}
	\item Being able to observe a significant difference and a certain similarity between two different executable files sharing a common sub-set of API calls but having a different overall behavior. This suggests that the individual API call features must be distributed across pixels, since our visual system is more sensitive to the overall texture of the image rather than tracking specific pixels. Also, since the tf-idf (of API calls or call sequences) features are generally sparse, any local representation would be very hard to visualize, as the bright pixels would also be sparse, with most of the image being dark.
	\item Encapsulating a long-range inter-dependence of the API calls across the execution, rather than just being limited by the ngrams (of API calls), such that, for example, each pixel intensity or a texture formed in the overall image representation is determined by every ngram -- no matter how much they are separated in the sequence of calls. 
	\item Being able to apply existing deep learning techniques which work efficiently on publicly available image datasets. This requires that the priors inherently assumed of the data, by such models, remain still  valid.  For instance, models designed for classification tasks involving regular images assume the requirement of translational invariance of the predictions with respect to the semantic content, i.e., for example, a picture of an object of interest  in a scene translated to a different location should still be classified as the same object. Under a local image representation of the file behavior, where a pixel position corresponds to an unique API call ngram,  a  convolutional filter map may  get activated by  similar pattern of pixel intensities  across different parts of an image under translation. This would lead to different behaviors getting similar activations, as positions encode unique sets of the API call ngrams in a local representation. Hence, for published image-based deep learning architectures to work, an image representation needs to have uniquely identifiable patterns for different events, across the image space.
	
\end{enumerate}

To achieve the above considerations, we exploit Fourier transform, in encoding the behavioral information into distributed image textures by mapping the frequency space with the individual API call ngram information. We elaborate the methodology of generating the image representations from the file behavior subsequently, after defining our notation for the Fourier Transform.

The Fourier Transform, $\widetilde{I}=\text{DFT}\{I\}$, over a single-channel real-valued Image ${I}$ of size $n \times n$ is defined by
\begin{equation}
\widetilde{I}[i][j] =\frac{1}{n^2} \sum^{n-1}_{a=0}~ \sum^{n-1}_{b=0} ~{I}[a][b] ~e^{-\iota2\pi\left(\frac{ia}{n}+\frac{jb}{n}\right)},
\label{dft}
\end{equation}
where the indexing notation $[p][q]$ refers to the value at the $p^{th}$ row and $q^{th}$ column. Note that $\iota$ above represents $\sqrt{-1}$. The Fourier transform of the image, $\widetilde{I}$, is a complex-valued $n \times n$ matrix, and therefore can be decomposed into the real and the imaginary parts, or equivalently, the amplitude and the phase parts, as illustrated below:
\begin{equation}
\label{equationI}
\widetilde{I}[i][j] \equiv \mathcal{A}[i][j] ~ e^{\iota \mathcal{P}[i][j]},
\end{equation}
where  $\mathcal{A}$ and $\mathcal{P}$ represent real-valued $n \times n$ sized matrices corresponding to the amplitude and the phase respectively with $\mathcal{A}[i][j]\geq0$ and $\mathcal{P}[i][j]\in[0, 2\pi)$. This operation can be implemented (per channel) using the OpenCV 2 package for Python with a few lines of code as follows:
\begin{tabbing}
\verb|amplitude, phase = cv2.cartToPolar(|\=\verb|cv2.dft(numpy.float64(image),| \\\>\verb|flags=cv2.DFT_SCALE| | \verb|cv2.DFT_COMPLEX_OUTPUT),|\\ \>\verb|angleInDegrees=True)|
\end{tabbing}

A location in the frequency space after DFT, corresponds to a periodic texture with a unique frequency and orientation  (refer to the DFT basis images, as an example, in Figure \ref{dftbasis}) in the position space, distributed across the pixels. We, therefore, encode the API call ngram words to the locations in the frequency space, with their tf-idfs determining the amplitudes in the amplitude spectrum $\mathcal{A}$. The phase spectrum $\mathcal{P}$ is exploited to encode the relative ranks of the API call ngrams in the order of their first-invocation. This is repeated across the three channels (R, G, B) independently corresponding to the 4-grams, 3-grams, and 1,2-grams, respectively. Finally, a distributed image representation is obtained by applying an inverse DFT (IDFT), given the amplitude and the phase spectra, defined over a single-channel matrix $\widetilde{I}$ (with $\widetilde{I}[i][j] \equiv \mathcal{A}[i][j] ~ e^{\iota \mathcal{P}[i][j]}$) of size $n \times n$   by
	\begin{align}
	I[a][b] &= \sum^{n-1}_{i=0} ~\sum^{n-1}_{j=0} ~\widetilde{I}[i][j] ~e^{\iota2\pi\left(\frac{ia}{n}+\frac{jb}{n}\right)} 
	= \sum^{n-1}_{i=0} ~\sum^{n-1}_{j=0} ~\mathcal{A}[i][j] ~e^{\iota2\pi\left(\frac{ia}{n}+\frac{jb}{n}\right)+ \iota \mathcal{P}[i][j]}.
	\label{idft}
	\end{align}
	Clearly, each  pixel  intensity in the position space depends on all the individual API call ngram features in the frequency space as evident from the above Eq. \eqref{idft}. This can be implemented (per channel) using OpenCV 2 as:
\begin{tabbing}
\verb|image = cv2.magnitude(|\=\verb|cv2.idft(|\verb|cv2.polarToCart(|\verb|amplitude, phase,|\\
\> \verb|angleInDegrees=True), |\verb|flags=cv2.DFT_COMPLEX_OUTPUT))|
\end{tabbing}

	When visualizing the file behavior as a RGB image by this method, not only do we have the API information per channel distributed across the image, forming textures, but also, the information across the channels are intermixing, to give a rich  color palette and unique patterns, to the final image for easy manual inspection and categorization as subsequently shown in Section \ref{sec:categorization}. 
	
Since the Fourier transform is reversible, one may also decode back the original API call ngrams with their first-invocation ranks and the term-frequencies, from the image representations losslessly.

We outline the algorithm for the image generation after describing the dataset below. 

\subsection{Dataset} Our dataset consisted of API call sequence logs  over $1,984$ distinct API functions, from a total of $12,006$ distinct PE (.EXE) files executed through RunningWater. RunningWater is a proprietary software from Symantec, for dynamic malware analysis, that hooks itself into the running process, capturing the API call events. The samples are executed on a VM, running Windows, on a single-core. The environment is restored back after each run, to avoid any interference across the samples. The sample binaries, for our research, have been downloaded  partly from the VirusTotal portal and partly from Symantec's own datastores. An example output of event sequence from RunningWater on execution of a file is as follows (ordered by the timestamp of invocation across threads)
\begin{tabbing}
$\cdots$\\
\verb|event(1501696951,8644,3120,api_GetEnvironmentVariable(_))| \\
\verb|event(1501696951,8644,3120,api_GetEnvironmentVariable(_))|\\
\verb|event(1501696951,8644,3120,api_RegQueryInfoKey(21900,0,0,_,5,9,0,0,0,0,0,0,0))|\\
\verb|event(1501696951,8644,3120,api_RegEnumKeyEx(21900,4,'v4.0',4,_,0,0,[3647740521,30361877]))|\\
$\cdots$
\end{tabbing}
 with the format: \verb|event(timestamp, process_id, thread_id, api_name(args))|. The run time per sample is capped at 2 minutes with sleep times (if any) limited to 1 second. The average number of events  across our  dataset was about 40,000 per file execution.

In this work,  we only consider the API calls as words, leaving out the arguments. One of the reasons being that, malwares, generally, tend to randomize the arguments wherever possible to evade string-based or hash-based static signatures. Therefore, from the  above shown example, we only consider  the following effective sequence when modeling a document representing the file behavior:
\begin{tabbing}
$\cdots$\\
\verb|GetEnvironmentVariable| \\
\verb|GetEnvironmentVariable|\\
\verb|RegQueryInfoKey|\\
\verb|RegEnumKeyEx|\\
$\cdots$
\end{tabbing}

We reserve including arguments and loaded DLL information into the sequence, for future work.
To assign labels (malicious or clean), we used Symantec's reputation service. Our final labeled dataset consisted of $1,662$ clean, $2,512$ malicious, and $7,832$ grayware (low confidence in assigning either label) files.

\clearpage
\subsection{Image Generation} The image representation of the API call sequence log is generated by an inverse DFT of the frequency space, where the tf-idfs of event ngrams (n=1 or 2, 3, and 4, mapped to the B, G, R channels, respectively) are encoded as the amplitude of a frequency component, and  the relative first-occurrence order of an event is preserved in the phase information. Here, we describe the algorithm for generating $64\times64\times3$ image representation from behaviors, that may be trivially extended to the general case of $n\times n\times3$ dimensions.
\begin{enumerate}
	\item Each API call sequence log, out of a total 12,006 (clean+malware+grayware) file execution logs,  is modeled as a document of words representing the  sequence of the API call invocations per file execution. 
	\item Based on this corpus of documents, a vocabulary consisting of 1, 2, 3, and 4-grams is built using the scikit-learn's \cite{scikit} TfidfVectorizer method as follows. 
	\begin{tabbing}
		\verb|sklearn.feature_extraction.text.TfidfVectorizer(|\=\verb|analyzer="word",|\\ \> \verb|max_df=0.98,|\\ \> \verb|min_df=1,| \\ \>\verb|ngram_range=(1, 4),| \\ \> \verb|norm=None,|\\ \> \verb|use_idf=True,|\\ \> \verb|smooth_idf=True,| \\ \>\verb|sublinear_tf=True)|
	\end{tabbing} 
	
	By having \verb|max_df=0.98|, we ignore the terms that occur in over 98\% of the documents considering them similar to stop words. No normalization was used to prevent the suppression of tf-idf terms across the (1,2)-grams, 3-grams, and 4-grams, as they are treated independently while mapping to the B, G, R channels of the image, respectively. Finally, the sublinear term-frequency \cite{stanfordtfidf}, generally defined as $1+\log($\verb|tf|$)$ where \verb|tf| is the non-zero term-frequency, was used in order to weigh down the effect of highly repetitive API call words as it is unlikely that twenty occurrences of a term in a document truly carries twenty times the significance of a single occurrence. Moreover, this also helps, upto some extent, against good-API call spamming, a trick generally used by the malware authors to evade detection. The $\log$ nonlinearity also helps in bringing the tf-idf features across documents to a comparable numerical range without skew.
	
	The vectorizer is fit over the corpus and transformed to generate the tf-idf vector representations for the documents in the corpus, before persisting it on disk for future use - during the prediction/evaluation phase. The vocabulary consisted of 876 unigrams, 61,052 bigrams, 365,388 trigrams, and 840,990 quadgrams after fitting it over our dataset.
	
	\item Since an image is finite in size (in this case, $64\times64\times3$), only a limited number of the ngrams may be encoded. Therefore, we choose only the top $64\times64=4096$ ngrams per channel that correlate the most with the ground-truth labels (clean or malicious) based on a holdout set. To prevent any leakage of the label information, we randomly choose 250 samples each out of a total 1,662 clean and 2,512 malicious samples, respectively, as holdout sets for calculating the significance of the ngrams, with the below definition, that intuitively represents the ratio of the inter-class distance to the intra-class deviation.
	\begin{align}
	\text{significance(}\text{i}\text{)}=\bigg|\frac{\mu^\text{i}_1-\mu^\text{i}_0}{\sigma^\text{i}_1+\sigma^\text{i}_0}\bigg| \label{significance}
	\end{align}
	Here, $\mu^\text{i}_j$ and $\sigma^\text{i}_j$ respectively represent the average/mean and the standard deviation of the tf-idf values corresponding to the ngram represented by i, among the samples belonging to the class $j$ ($j=$1 for malicious, 0 for clean).
	
	The top 20 ngrams across channels with the highest significance measure were as shown in the Table \ref{table1}.

\begin{table}
  \caption{Top 20 discriminating ngrams with their Inverse Document Frequencies (IDF).}
  \label{table1}
  \centering
  \begin{tabular}{lll}
    \toprule
    ngram (n=1,2,3,4)     & Significance     & IDF \\
    \midrule
    \verb|NtClose NtOpenKey NtQueryValueKey NtClose|  & 0.602208 & 4.317899     \\
  \verb|ProcessIdToSessionId|     &0.593848 & 3.836467     \\
    \verb|RtlDeleteBoundaryDescriptor|     & 0.571707      & 4.192736  \\
\verb|NtQueryValueKey NtCreateFile|     & 0.524891      & 4.634343 \\
\verb|NtQueryValueKey NtCreateFile NtCreateSection|     & 0.524891       & 4.634343  \\
\verb|NtQueryValueKey NtCreateFile NtCreateSection NtMapViewOfSection|     & 0.524891      &4.634343  \\
\verb|VirtualAlloc VirtualAlloc VirtualAlloc VirtualProtect|     & 0.513892       & 2.501540  \\
\verb|GetStartupInfo|     & 0.508476      & 4.421983  \\
\verb|NtClose NtQueryValueKey|     & 0.508327       & 4.283998  \\
\verb|NtOpenProcessToken NtQueryInformationToken NtClose NtOpenKey|     & 0.506242       & 4.643852  \\
\verb|NtQueryInformationToken NtClose NtOpenKey NtQueryValueKey|     & 0.500000       & 4.686135  \\
\verb|RtlDeleteBoundaryDescriptor RtlDeleteBoundaryDescriptor|     & 0.497304      & 4.343512  \\
\verb|NtOpenThreadToken|     & 0.491294       & 4.477043  \\
\verb|RtlGetNtProductType RtlDeleteBoundaryDescriptor|     & 0.478606       & 4.682818  \\
\verb|RtlDeleteBoundaryDescriptor RegOpenKeyEx|     & 0.476774       & 4.669660  \\
\verb|QueryPerformanceCounter GetStartupInfo|     & 0.474858      & 4.603285  \\
\verb|GetSystemTimeAsFileTime QueryPerformanceCounter GetStartupInfo|     &0.474858      & 4.612502  \\
\verb|NtQueryValueKey|     & 0.472702       & 1.707998  \\
\verb|Process32Next|     & 0.469005       & 3.009256  \\
\verb|LoadLibrary ZwQuerySystemInformation|     & 0.466916       & 2.442038  \\

    \bottomrule
  \end{tabular}
\end{table}
	
	\item Top $n\times n$ (here $64\times64=4096)$ ngrams for each of the three independent channels, i.e., 61,928 1,2-grams, 365,388 3-grams, and 840,990 4-grams, are picked based on the significance measure defined in Eq. \eqref{significance} calculated over the holdout set of 500 images (250 clean, and 250 malicious) as described above. These 500 images are  completely removed from the dataset  to prevent any information leak biasing our results. The rest of the steps are performed over the remaining 1,412 clean and 2,262 malicious files.

  \begin{figure}[tp]
\centering
\includegraphics[scale=0.3]{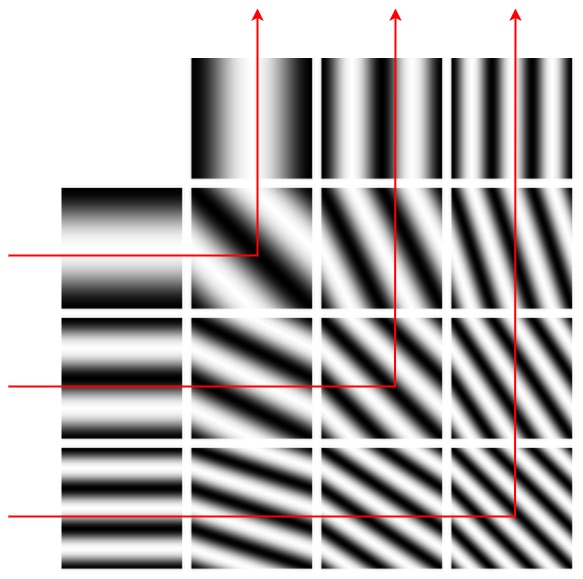}	
\caption{DFT basis images in the position space corresponding to an amplitude for a single frequency out of $4\times4=16$ possible frequencies with varying orientation where the top-left pixel represents the DC component that determines the overall brightness of the image in the position space. The arrows represent the order of assigning the ngrams based on their significance starting from the bottom-most arrow.}
\label{dftbasis}
\end{figure}
	
	\item Across each channel, the higher significant ngrams are mapped to positions of higher frequency as occupied when transforming an image into frequency-space by DFT. This allows one to model better discriminating features with sharper distinctive edges into the position-space. This methodology is chosen such that the image representation is visually differentiable, for manually categorizing files.  The lesser discriminative features are assigned to the  lower frequency positions in the Fourier-space of the image, which, effectively, get modeled as smoother patterns in the position-space image representation. An example of the  DFT basis images for the case of $4\times4$ dimensions are shown in Figure \ref{dftbasis}, along with a demonstration of the ngram to coordinate mapping (in the frequency-space) in Table \ref{table2} for  the Blue channel (n=1 or 2). This is repeated similarly for the Red and Green channels, corresponding to the 4-grams and 3-grams, respectively. The top left-most pixel in the frequency space corresponds to the DC component,  determining the overall brightness of the image, that gets scaled appropriately on normalizing the contrast per channel (described subsequently). 

\begin{table}
  \caption{Top 15 1,2-grams assigned to the positions in the frequency space of the Blue channel of a $4\times4$ Image. The height $h$ (along the vertical axis) and the width $w$ (along the horizontal axis) coordinates are measured from the top-left position labeled by $(0,0)$. }
  \label{table2}
  \centering
  \begin{tabular}{lll}
    \toprule
    ngram (n=1,2)     & Significance     &  Position \\ &&assigned $(h, w)$  \\
    \midrule
    \verb|ProcessIdToSessionId|  & 0.593848 &  $(3,0)$     \\
  \verb|RtlDeleteBoundaryDescriptor|     &0.571707 & $(3,1)$     \\
    \verb|NtQueryValueKey NtCreateFile|     & 0.524891      & $(3,2)$  \\
\verb|GetStartupInfo|     & 0.508476      & $(3,3)$ \\
\verb|NtClose NtQueryValueKey|     &0.508327       & $(2,3)$ \\
\verb|RtlDeleteBoundaryDescriptor RtlDeleteBoundaryDescriptor|     & 0.497304     &$(1,3)$ \\
\verb|NtOpenThreadToken|     & 0.491294       &  $(0,3)$   \\
\verb|RtlGetNtProductType RtlDeleteBoundaryDescriptor|     & 0.478606      & $(2,0)$  \\
\verb|RtlDeleteBoundaryDescriptor RegOpenKeyEx|     & 0.476774       & $(2,1)$  \\
\verb|QueryPerformanceCounter GetStartupInfo|     & 0.474858       & $(2,2)$  \\
\verb|NtQueryValueKey|     & 0.472702       & $(1,2)$  \\
\verb|Process32Next|     & 0.469005      & $(0,2)$  \\
\verb|LoadLibrary ZwQuerySystemInformation|     & 0.466916       & $(1,0)$\\
\verb|CreateToolhelp32Snapshot|     & 0.461527       & $(1,1)$  \\
\verb|NtOpenKey NtQueryValueKey|     & 0.451770       & $(0,1)$  \\
DC Component (initialized with a positive value) that gets &&\\  scaled on normalizing the contrast    & --    & $(0,0)$  \\
    \bottomrule
  \end{tabular}
\end{table}

\item For each of the ngrams mapped to a position in the frequency space per channel, the corresponding tf-idf coefficients are encoded as the  amplitudes in the polar representation of the DFT. We denote this by the matrix $\mathcal{A}^j_i$, given an executable $i$, across the channels denoted by $j\in\{\text{R, G, B}\}$.

\item To determine the phase component of the call ngrams mapped to the positions in the frequency space,  the relative first-invocation ranks  of the ngrams (i.e., in the sequence of the ngrams ordered by their first-invocation in the API call log) per channel are computed for a given executable. That is, considering the R channel, for instance, the relative ranks of  first-invocation are computed among the 4-grams that occurred in the executable. The ranks assigned per channel are  then rescaled to $(0, 360]$, dividing the range into equal intervals, to encode the phase corresponding to the ngram positions in the frequency space. We denote this by the matrix $\mathcal{P}^j_i$  for a given executable $i$ across the channels denoted by $j\in\{\text{R, G, B}\}$. Therefore, the sequence information is not only captured with the tf-idfs of the ngrams (n>1), but also into the phase of the image, by encoding the   relative first-occurrence ranks among the ngrams per channel.

For the case of the Blue channel consisting of both 1 and 2-grams, the ngram ranks assigned among  each set are merged in the same order, giving a higher precedence to the 1-grams over the 2-grams in the case of a tie. As an example, consider the sequence \verb|Q, Q, B, P|. The ranks for the 1-grams would be 1, 2, 3 corresponding to \verb|Q, B, P|; while for the 2-grams, the terms \verb|QQ, QB, BP| would carry the ranks $1, 2, 3$, respectively, by our method. The final ranks are then determined to be $1, 2, 3, 4, 5, 6$ for the terms in the order \verb|Q, QQ, B, QB, P, BP|, respectively. The ranks  rescaled to degrees correspond to $60^\circ, 120^\circ, 180^\circ, 240^\circ, 300^\circ, 360^\circ$ phases, respectively, with a default value of $0^\circ$ for terms that do not appear in the execution. Similarly, the phases computed for the Green channel would  be $180^\circ$ and $360^\circ$ corresponding to the 3-grams \verb|QQB| and \verb|QBP|, respectively, by the same procedure. 

\item Given the assigned phases and the amplitudes for an  executable $i$,  the position space image for the channel $j$ is computed as
\begin{align*}
\text{Image}^j_i = \text{ContrastNormalize}(\text{IDFT}\{\mathcal{A}^j_i, \mathcal{P}^j_i\}),
\end{align*}
where ContrastNormalize($\cdot$) linearly rescales the pixel value range per channel after IDFT to $[0, 255]$, that fixes the amplitude of the DC component in the frequency space. Finally, the RGB color image representation  of  the executable $i$ is obtained by stacking the IDFTs along the depth as 
\begin{align*}
\text{Image}_i = [\text{Image}^R_i, \text{Image}^G_i, \text{Image}^B_i].
\end{align*}
This is stored as a PNG to avoid any loss due to image compressions  inherent to alternative file formats such as the JPEG. Figure \ref{figencoding} provides an illustration of the image encoding scheme described above.
\begin{figure}[tp]
	\centering
\makebox[\textwidth][c]{\includegraphics[width=1.4\textwidth]{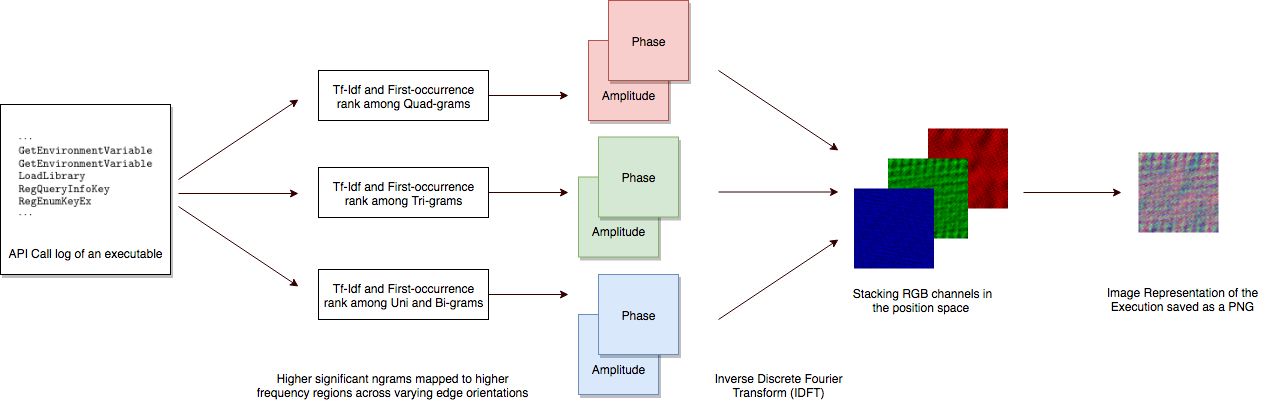}}%
		\caption{An overview of the image encoding scheme}
	\label{figencoding}
\end{figure}
\end{enumerate}

\begin{figure}[tp]
	\centering
	\subfloat[Non-malicious (clean) files]{
		\includegraphics[scale=0.4]{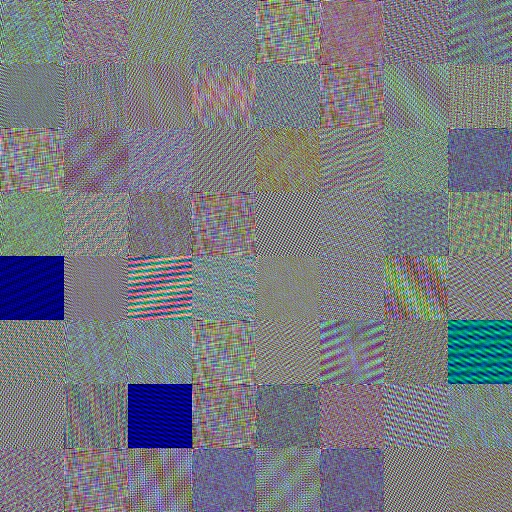}
		
	}
	\centering
	\subfloat[Malicious files]{
		\includegraphics[scale=0.4]{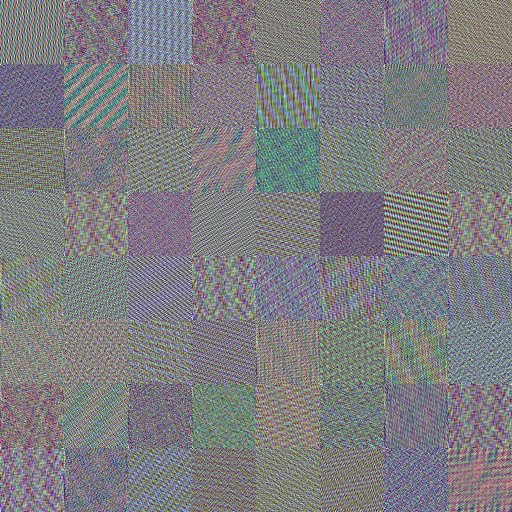}
		
	}
	\caption{Samples of $64\times 64$ image representation corresponding to $64$ distinct files per category randomly chosen from the dataset. The R, G, B channels represent the 4-gram, 3-gram, and 1,2-grams respectively of the API call words. The amplitude per pixel after DFT is determined by the corresponding tf-idf, and the phase by its relative first-occurrence rank among its channel of ngrams. (Images shown are scaled.) \label{dataset}
	} 
	\label{images1}
\end{figure}

Figure \ref{images1} shows 64 samples of $64\times64$ color images generated across the clean and malicious set of files with the steps described above. Each image visually encodes a total of $(4096-1)\times 3= 12,285$ relative tf-idf coefficients of the ngrams across channels, with the first-invocation order captured in the phase of the image. The position space image representation corresponds to a weighted (by tf-idf coefficients) sum of 2D sinusoidal waves at different frequencies, orientations (corresponding to different ngrams), and  phases (corresponding to the first-occurrence order). In addition to this, the color channels (Blue, Green, or Red) of the image, representing the ngram type, i.e., $n=1$ or $2, 3,$ and $4$, combine across the depth, giving rise to a spectrum of composite colors for a rich image visualization of the file behavior.

\subsection{Decoding Images back to Sequence} \label{sec:decoding} 
\begin{figure}[tp]
	\centering
\makebox[\textwidth][c]{\includegraphics[width=1.4\textwidth]{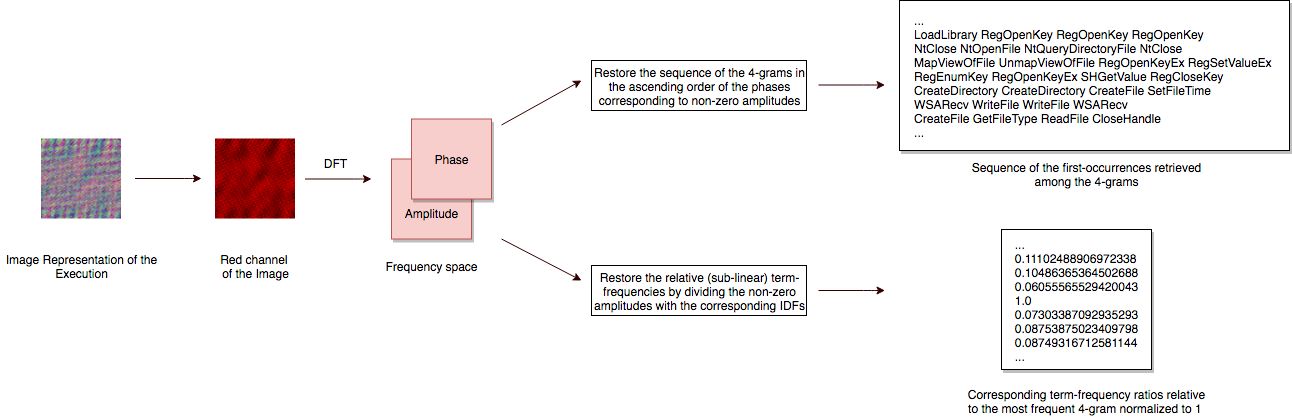}}%
		\caption{An overview of the image decoding scheme per channel}
	\label{decodingimage}
\end{figure}

Since the Fourier Transform is reversible, one can decode back the information encoded into the images losslessly. The phase of the image at a position in the frequency space can be used to rank the corresponding ngram word to determine its relative order in the output ngram sequence, and the amplitude can be used to derive the relative tf-idf value. The IDF coefficients computed for each ngram word previously, during image generation, can be used to extract the relative (sub-linear) term-frequencies of the ngrams from their relative tf-idfs (amplitudes). This is summarized in the Figure \ref{decodingimage} for the R channel that corresponds to the 4-grams here.  The same procedure may be repeated for the other channels to decode the tri- and bi/uni- grams respectively. Being able to decode the images allows interpretability of the decisions made by machine learning classifiers built over the image representations, helping manual analysis of critical false-positives or false-negatives.

\section{Applications on Software/Malware Categorization and Smart-Definitions} \label{sec:categorization} The described image representation of a file behavior, summarizes the API content and the sequence information into the visible artifacts, such as  edges at different orientations, their relative intensities, and the color of the pixels. This enables easy  software and malware categorization, manually, by recognizing the similarities and the differences among the executables, based on the visual patterns. Quick manual malware analysis is critical, especially, for zero-day outbreaks, such as the recent WannaCry attack, to reduce the damaging impact. For such cases, the image representation described, may accelerate the analysis through easy visual comparison with other known malwares.  Attribution to  related malwares, for instance, can help in faster remediation. It also offers the flexibility of using color filters (Red/Green/Blue) for comparing patterns channel-wise, or even applying band-pass filter operations instead, to only compare the patterns formed by a particular set of ngrams across the given samples. 

Categorization may also be easily automated using general image similarity methods and clustering techniques. Since the images  have highly ordered structure, unlike the pictures of objects or a scene in real-life, simple distance metrics, such as the pixel-wise Euclidean distance, may be employed for clustering the files into categories. Other methods that summarize the perceptual information or the semantic content of the image (for instance, based on the pre-final layers of a trained deep convolutional neural network) may also be used for image similarity and clustering.

Perceptual  hashing techniques  (such as the dHash \cite{dHash}, and pHash  \cite{pHash})  applied over the image representation, have an advantage in generalizing well across variants of files with similar behavior, in contrast to the cryptographic hashes (like the SHA256)  that are more commonly  used for fingerprinting files  as anti-malware definitions. Cryptographic hashes change drastically with slight modification in the content, and therefore, carry no meaningful distance metric for measuring similarity. Storing the perceptual hashes of the image representations as \textit{smart-definitions} instead of the SHA256-based hashes, for instance, can improve performance, generalization, and also reduce disk footprint of the antivirus softwares.  

Table \ref{categorization} demonstrates the categorization of softwares/malwares based on the visual similarity of our proposed image representations, while comparing a distance metric based on the  dHash \cite{dHash}. Note that each of the files and the images displayed are different (i.e., have unique SHA256 digests) although some images or the file names, may look similar.

\begin{figure}[tp]
	\centering
	\includegraphics[scale=0.6]{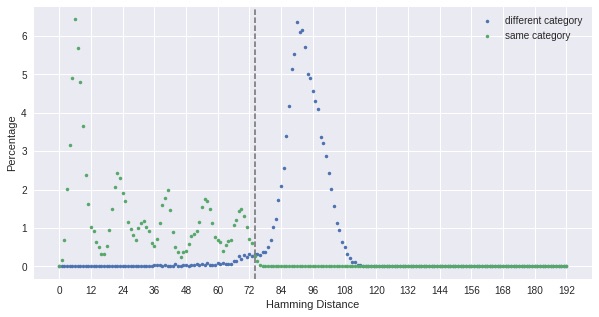}
	\caption{Normalized histogram (bin width = 1) of pair-wise Hamming distances based on a 192-bit perceptual hash, obtained by concatenating the 64-bit dHashes \cite{dHash} computed per channel, across 254 samples belonging to 21 file categories determined from  Symantec's internal metadata combined with VirusTotal based on file paths, virus names, etc., in-field across clients.}
	\label{phashcutoff}
\end{figure}

The  generalization capability of dHash, for instance, which is based on the pixel gradients, can be evidently noted  from the Figure \ref{phashcutoff}. This clearly shows a separation of the pair-wise Hamming distances  among the files of the same vs different categories with the grey line representing our proposed cutoff of $74$. Stronger cutoffs on the Hamming distance may be required for FP-sensitive applications such as malware classification using the perceptual hash-based similarity.

\begingroup
\setlength{\LTleft}{-20cm plus -1fill}
\setlength{\LTright}{\LTleft}
\small
\begin{longtable}{lllll}
 \caption{Software categorization based on the images. Note that the images and the files tabulated below are all distinct with different SHA256 digests. The Hamming distances are computed with respect to the first row of each category based on a 192-bit perceptual hash obtained by concatenating the 64-bit dHashes \cite{dHash} computed per channel. Last three categories demonstrate malware categorization.}
  \label{categorization}
  \\

\toprule \\Category   & Image (scaled)    &  Filename& Difference Hash or dHash (192-bit Hex) & Hamming \\ &&&&Distance \\ \hline \hline
\endfirsthead

\endhead
\\
\\ \multicolumn{5}{r}{{Continued on next page.}} \\
\endfoot

\hline \hline
\endlastfoot
\\

Adobe Acrobat Reader & \parbox[c]{1em}{\includegraphics[width=0.5in]{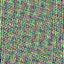}}  \vspace{0.06cm}       & acrord32-911.exe   & \begingroup \scriptsize \texttt{631846328c6399ce2a524242229181d5ab5ad52a54ad6a55} \endgroup & 0\\
 & \parbox[c]{1em}{ \includegraphics[width=0.5in]{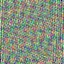}} \vspace{0.06cm}    & acrord32-950.exe   & \begingroup \scriptsize \texttt{e1810e38e0c31f4c2a42004042a081d5ab5ad52a56ad6a55} \endgroup & 31\\
 & \parbox[c]{1em}{ \includegraphics[width=0.5in]{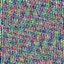}} \vspace{0.06cm}      & acrord32-992.exe   & \begingroup \scriptsize \texttt{33c1263ee0230fcc0a1e704006bcf9c1cb4a6d2d34b592d2} \endgroup & 64\\
 & \parbox[c]{1em}{ \includegraphics[width=0.5in]{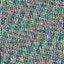}}\vspace{0.06cm}       & acrord32-918.exe   & \begingroup \scriptsize \texttt{33c13e3ee0030fec0a3e784106bdf1c1cb4a6d2d34b59252}\endgroup & 68\\
\midrule
MySQL Database & \parbox[c]{1em}{\includegraphics[width=0.5in]{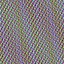}}  \vspace{0.06cm}       & mysqladmin.exe (5.1 MB)   & \begingroup \scriptsize \texttt{00010000000000810f0783e1f0783c1ec9a92a221656d5d5}\endgroup & 0\\
 & \parbox[c]{1em}{ \includegraphics[width=0.5in]{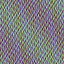}} \vspace{0.06cm}    & mysql\_embedded.exe (24.1 MB)   & \begingroup \scriptsize \texttt{81800040000000810f0783c1e0f87c3ead2dad2ab2525559}\endgroup & 28\\
 & \parbox[c]{1em}{ \includegraphics[width=0.5in]{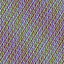}} \vspace{0.06cm}    & mysqld.exe (37.9 MB)   & \begingroup \scriptsize \texttt{00000000000000800f07c3e1f0783c1eadad2ca21252d355}\endgroup & 15\\
\midrule
TeX Typesetting System & \parbox[c]{1em}{\includegraphics[width=0.5in]{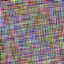}}  \vspace{0.06cm}       & miktex-luatex.exe   & \begingroup \scriptsize \texttt{a56c4a8b9b92a3ade952f5ac54b13ad42db6464452592cb8}\endgroup & 0\\
 & \parbox[c]{1em}{ \includegraphics[width=0.5in]{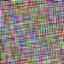}} \vspace{0.06cm}    & miktex-xetex.exe   & \begingroup \scriptsize \texttt{a56c4a8b9b92a3ad697371ac4eb13a942da646445269acb8}\endgroup & 13\\
 & \parbox[c]{1em}{ \includegraphics[width=0.5in]{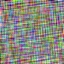}} \vspace{0.06cm}    & bg5pdflatex.exe   & \begingroup \scriptsize \texttt{256c4a8b8b92a9ade95371ac4eb5329429a64644524368b8}\endgroup & 20\\
 & \parbox[c]{1em}{ \includegraphics[width=0.5in]{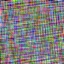}} \vspace{0.06cm}    & miktex-bibtex.exe   & \begingroup \scriptsize \texttt{256c4a8b8b92a92de95371ac46b13a942da64644526328b8}\endgroup & 17\\
\midrule
Norton Download Managers & \parbox[c]{1em}{\includegraphics[width=0.5in]{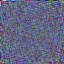}}  \vspace{0.06cm}       & norton\_download\_manager-1.exe   & \begingroup \scriptsize \texttt{31899d0a24d883cc8c3698d95a1965349130056d56960de8}\endgroup & 0\\
 & \parbox[c]{1em}{ \includegraphics[width=0.5in]{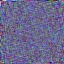}} \vspace{0.06cm}    & nortonn360downloader.exe   & \begingroup \scriptsize \texttt{31899d4a24d883cc4c3698895a196519b130356dd2d04d6c}\endgroup & 20\\
\midrule 
Bluestacks Android Emulator & \parbox[c]{1em}{\includegraphics[width=0.5in]{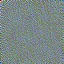}}  \vspace{0.06cm}       & Bluestacks.exe   & \begingroup \scriptsize \texttt{282c2b0aaa1a22b223c3d80d652598908655c4526a782ca9}\endgroup & 0\\
& \parbox[c]{1em}{ \includegraphics[width=0.5in]{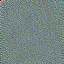}} \vspace{0.06cm}    & BlueStacksTV.exe   & \begingroup \scriptsize \texttt{282c2b09aa1b20a323c3c80d65659c910655d452e878aca9}\endgroup & 15\\
\midrule
Installers & \parbox[c]{1em}{\includegraphics[width=0.5in]{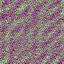}}  \vspace{0.06cm}       & installer\_solmysterystolenpower\_es.exe   & \begingroup \scriptsize \texttt{0280009100000088a545cd99b92a6964926d920db28512ac}\endgroup & 0\\
 & \parbox[c]{1em}{ \includegraphics[width=0.5in]{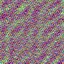}} \vspace{0.06cm}    & installer\_governorofpokerdeluxe\_de.exe   & \begingroup \scriptsize \texttt{0080009000000088a545cd99b92a6964926d920db285122c}\endgroup & 3\\
 \midrule
Google Updater & \parbox[c]{1em}{ \includegraphics[width=0.5in]{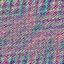}} \vspace{0.06cm}      & googleupdate.exe   & \begingroup \scriptsize \texttt{52b2f645899193130220000000408174d4a9d52a54a8552d}\endgroup & 0\\
 & \parbox[c]{1em}{ \includegraphics[width=0.5in]{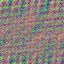}}   \vspace{0.06cm}    & googleupdate.exe   & \begingroup \scriptsize \texttt{52b2f62589931313328000800040104254a9d52a54a855af}\endgroup & 19\\
 \midrule
Windows Updates & \parbox[c]{1em}{ \includegraphics[width=0.5in]{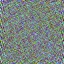}} \vspace{0.06cm}      & WindowsServer2003-KB958655.exe   & \begingroup \scriptsize \texttt{2c2c2622d2d2d999932a4cb64b53a4690603d13578ba2c8f}\endgroup & 0\\
 & \parbox[c]{1em}{ \includegraphics[width=0.5in]{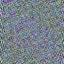}}   \vspace{0.06cm}    & WindowsInstaller-KB893803-v2-x86.exe   & \begingroup \scriptsize \texttt{2d2c2612d0d1d909932a4cb26b53a4680627c11178fa3c0f}\endgroup & 19\\
 \midrule
Adware.Graftor & \parbox[c]{1em}{ \includegraphics[width=0.5in]{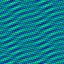}} \vspace{0.06cm}      & lollipop\_12062228.exe   & \begingroup \scriptsize \texttt{000000000000000048800000000040925200000181011124}\endgroup & 0\\
 & \parbox[c]{1em}{ \includegraphics[width=0.5in]{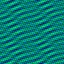}}   \vspace{0.06cm}    & lollipop\_12140006.exe   & \begingroup \scriptsize \texttt{000000000000000048800000000040925a04020981050224}\endgroup & 8\\
 \midrule
Backdoor:Win32/Cycbot.G & \parbox[c]{1em}{ \includegraphics[width=0.5in]{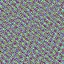}} \vspace{0.06cm}      & asedsle7   & \begingroup \scriptsize \texttt{009209910241a0082a810080008100c21a54e0021eb4e183}\endgroup & 0\\
 & \parbox[c]{1em}{ \includegraphics[width=0.5in]{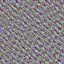}}   \vspace{0.06cm}    &C29.exe   & \begingroup \scriptsize \texttt{842049a440008249aa01000100810051863cf0020f3cf1c1}\endgroup & 42\\
 & \parbox[c]{1em}{ \includegraphics[width=0.5in]{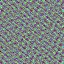}}   \vspace{0.06cm}    & DPYRAMYMEG-447.pms.exe.SVD   & \begingroup \scriptsize \texttt{01042182244992012a810001018100d11a54a10a1ed5a183}\endgroup & 32\\
 \midrule
Babylon PUA/Adware & \parbox[c]{1em}{ \includegraphics[width=0.5in]{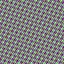}} \vspace{0.06cm}      & MyBabylonTB.exe   & \begingroup \scriptsize \texttt{0000000000000000a8810081008100d5aa00000000809085}\endgroup & 0\\
 & \parbox[c]{1em}{ \includegraphics[width=0.5in]{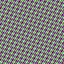}}   \vspace{0.06cm}    & Babylon.exe   & \begingroup \scriptsize \texttt{0000000000000000a1008100810081528a00000000800085}\endgroup & 21\\

\end{longtable}

\endgroup 

Recent works on data-driven perceptual hashing algorithms  involving deep convolutional siamese networks with binary output neurons \cite{deephash1,deephash2,deephash3,deephash4,deephash5}, that are tailored to the domain of application, promise further generalization of the hash fingerprints at lower FP rates.

\section{Simple Malware Classifier trained on the image pixels} \label{sec:classifier}
In this section we validate our proposed scheme of mapping the file behavior to the RGB image space by investigating the efficacy of a simple boosted-tree based model (using the XGBoost package \cite{xgboost}) for  malware classification. The model is trained directly over the flattened raw pixels of the  image representation corresponding to the 1,412 clean and 2,262 malicious samples. The model is trained to minimize the logloss cost function. The hyperparameters \daggerfootnote{\url{https://xgboost.readthedocs.io/en/latest/python/python_api.html}}  were chosen over a randomized grid search evaluated by the out-of-fold (OOF) validation AUC score.

\begin{table}
  \caption{5-Fold cross-validation  metrics, and confusion matrices   across different FPRs over the OOF predictions of the trained model.}
  \label{xgb64table}
  \centering
  \begin{tabular}{ll}
  \begin{tabular}{lllll}
    \toprule
   & \multicolumn{2}{c}{Validation} & \multicolumn{2}{c}{Training}\\
    \cmidrule(r){2-3} 
    \cmidrule(r){4-5} 
  CV Fold   & AUC     & Logloss     & AUC & Logloss \\
    \midrule
 fold1 & 0.9712     & 0.3038     & 0.99 & 0.0036 \\
 fold2 & 0.9754     & 0.2889     & 0.99 & 0.0038 \\
 fold3 & 0.9716     & 0.3205     & 0.99 & 0.0042 \\
 fold4 & 0.9743     & 0.3036     & 0.99 & 0.0040 \\
 fold5 & 0.9720     & 0.2948     & 0.99 & 0.0050 \\
\cmidrule(r){1-3}
all OOFs & \textbf{0.9727} & 0.3023 & & \\
    \bottomrule
  \end{tabular}
&
    \begin{tabular}{lllllll}
    \toprule
     TN     &  FN & TP & FP & TPR\% & FPR\%\\
    \midrule
   1411 & 1125  & \textbf{1137} & \textbf{1}  &\textbf{50.27} &\textbf{0.0708}     \\
    1405 & 687  & 1575 & 7  &69.63 &0.496     \\
    \bottomrule
  \end{tabular}
    \end{tabular}
\end{table}

\begin{figure}[tp]
	\centering
		\includegraphics[scale=0.6]{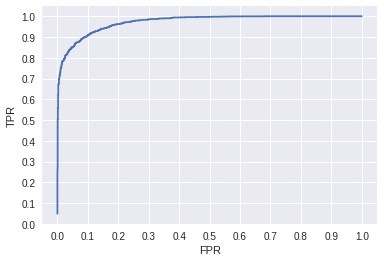}
		\caption{ROC curve over the OOF predictions of the  trained XGBoost model.}
	\label{rocplot}
\end{figure}

The  OOF cross-validation scores, ROC curve, and the confusion matrices at different FPRs (where `positive' refers to a conviction and `negative' to an exoneration) of the trained model  are summarized in the Table \ref{xgb64table} and Figure \ref{rocplot}.  The model achieves a TPR of about 50\% at a low FP rate of 0.071\% (corresponding to a single FP). The results show that the raw pixels of our image representation  encode significant information across the API call sequence  to discriminate between the two classes (malicious or benign) efficiently.

\section{Modeling Malware Behaviors with Generative Adversarial Networks}
\label{sec:gan}
Generative Adversarial Networks or GANs, first invented by Goodfellow \textit{et al.} \cite{goodfellow1} in 2014, are a class of generative models that approximate the inherent distribution of the input data, based on a finite sample represented by the training dataset,  using backpropagation over a minimax objective function involving two competing sub-networks — the discriminator and the generator — playing an adversarial zero-sum game, in an unsupervised setting, iteratively, where the generator minimizes the predictability of the discriminator in classifying the real  from the \textit{synthetic} data that is generated by the generator, and, in turn, the discriminator  competes by learning on the differences.

The weights of the discriminator network are updated   to minimize the loss in the classification of the fake (generated) from the real samples, whereas,  the generator is updated, backpropagating the discriminator's error signal, to maximize the classification loss of the discriminator, by generating fake samples which look more  real. Soon the generator is able to sample images highly resembling the input data while the discriminator saturates to being highly uncertain in classifying them from the real ones. The trained generator $G$, capturing the true input distribution, can then be sampled to interpolate or extrapolate the training data.

GANs, as originally proposed by Goodfellow \textit{et al.}, were limited by major challenges such as  training instability, high sensitivity to network hyperparameters, difficulty in choosing the right learning rates for the discriminator and the generator (so that neither saturates too fast), in addition to not being inherently convolutional for modeling image datasets. In 2015, Radford \textit{et al.} \cite{chintala} introduced a deep and convolutional version of the GANs called DC-GANs with improved training stability for modeling the image datasets, generating realistic synthetic images,  with interesting vector arithmetic in the input noise space that samples the learnt distribution of the generator. However, it still suffered from problems such as uninterpretable loss function that didn't always correlate  with the quality of the generated images, sensitivity to network architecture and hyperparameters, etc. 

In 2017, Arjovsky \textit{et al.} \cite{wgan1} proposed an alternative cost function for the GAN that estimates the Wasserstein distance of the synthetic to the real data distribution under the constraint of weight clipping, with the discriminator (or the ``critic'') predicting a value function instead of the class labels. They showed that, unlike the previous works, the loss function correlated well with the output quality of the generated images, in addition to an improved stability of the training. However, using weight-clipping strongly regularized the model with a high sensitivity to the clipping parameter $c$, limiting its capacity in capturing the higher moments of the data distribution, as shown by Gulrajani \textit{et al.} in \cite{wgan2} who suggested penalizing the gradient norm instead of clipping the weights. Their model, referred to as WGAN-GP, drastically improved the training stability across a wide variety of network architectures, notably also involving also the residual connections \cite{resnet}, with minimal or no tuning of the GAN hyperparameters. In addition, the modified  cost function of the GAN also ensured that the generator continued improving significantly  even after the discriminator neared saturation.

\subsection{Training}
We trained a WGAN-GP model over the image representation of the malicious file behaviors that inherently captures the API content along with the sequence of the first-occurrence. Once the generator is trained to model the distribution of the malicious behavior, one can sample instances that represent new malware, thereby, emulating a malware author against which a malware classifier may be tuned, tested, or trained for proactive protection.

\begin{figure}[tp]
	\centering
	\subfloat[Sample of the images used for training. ]{
		\includegraphics[scale=0.4]{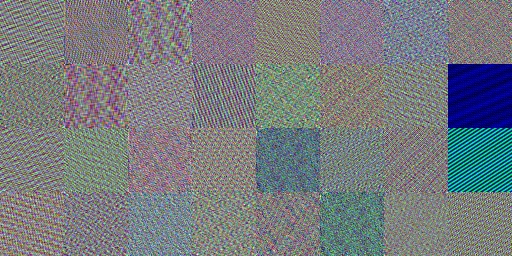}
		
	} 
	\centering
	\subfloat[Sample of the images used for validation.]{
		\includegraphics[scale=0.4]{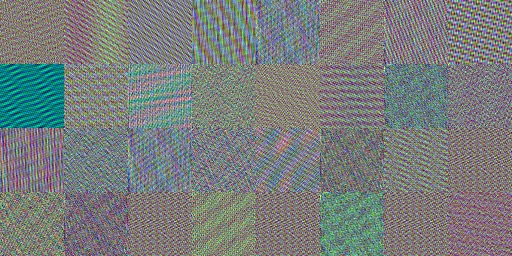}

	}
	\caption{Sample of the image representations of malicious files used for training the WGAN-GP model.
	} 
	\label{gandatasample}
\end{figure}

\begin{figure}[tp]
	\centering
\makebox[\textwidth][c]{\includegraphics[width=1.5\textwidth]{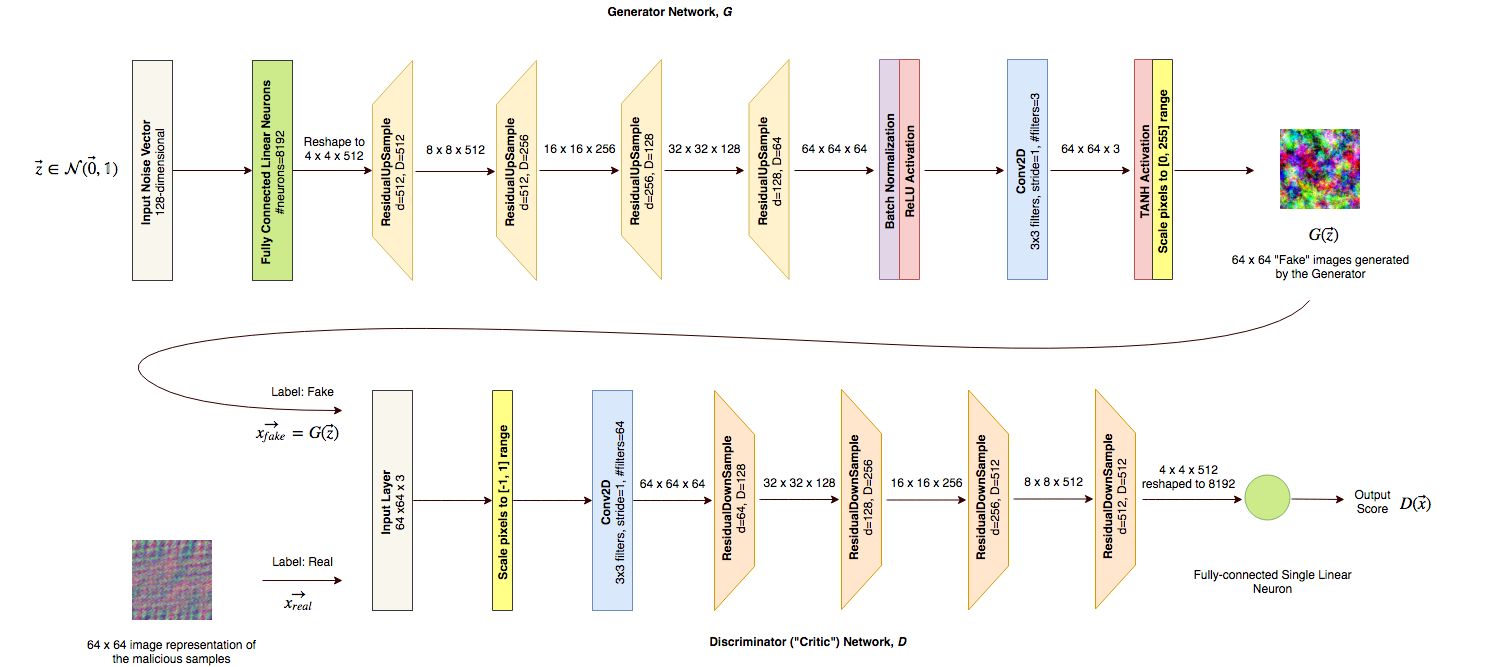}}%
		\caption{Architecture of the generator and the critic networks of the WGAN-GP model trained where the \textit{ResidualUpSample} and the \textit{ResidualDownSample} blocks are as defined in the Figures \ref{residualup} and \ref{residualdown}, respectively. }
	\label{fig:arch}
\end{figure}

Our dataset consisted of $64\times64\times3$ image representations of $2,262$  distinct malicious samples (excluding the $250$ malicious samples out of the $500$ used for mapping the top significant ngrams to the image frequencies). We divided the set into a 80:20 train:test holdout split to train the WGAN-GP model on $1,809$ and validate on $453$ image representations of distinct malicious samples. A sample of the training and the validation images used are shown in the Figure \ref{gandatasample}. 
The architecture of the generator and the critic network of the trained WGAN-GP model is summarized in the Figure \ref{fig:arch}, that has been one of the many architectures validated by Gulrajani \textit{et al.} in \cite{wgan2} who demonstrated competitive results of sample generation quality on the $128\times128\times3$ LSUN bedrooms dataset \cite{lsunbedrooms}. 

\begin{figure}[tp]
	\centering
	\subfloat[]{
		\includegraphics[scale=0.3]{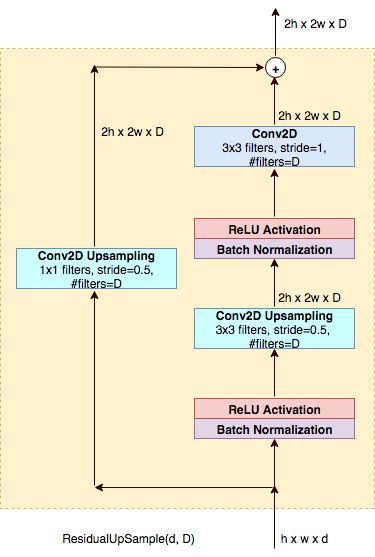}
		\label{residualup}
	} 
	\centering
	\subfloat[]{
		\includegraphics[scale=0.3]{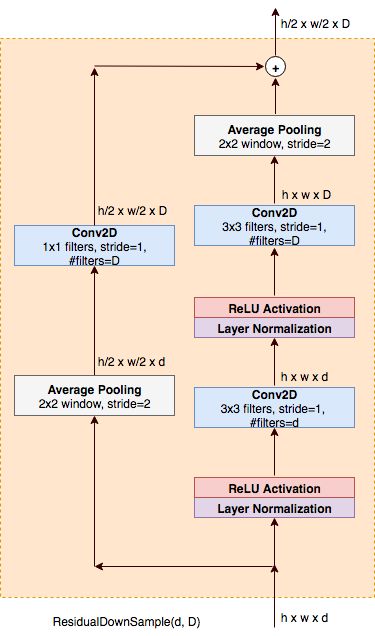}
		\label{residualdown}
	}
	\caption{Architecture of the residual blocks used in the (a) generator, and the (b) critic networks, respectively.
	} 
	\label{fig:residual}
\end{figure}
The critic:generator training iteration ratio is kept at $5:1$, i.e., for every $5$ training iterations of the critic, the generator is updated once. A default value of $\lambda=10$, as suggested in \cite{wgan2}, is used as the gradient penalty hyperparameter in the WGAN-GP loss function. A batchsize of $64$ is used for computing the gradient update per iteration that is parallelized into batches of $16$ training samples (data parallelisation) across $4$ nVIDIA GTX TITAN X GPUs with $12$ GB of available memory and $3,072$ CUDA cores each. Adam \cite{adam} optimization algorithm was used for training the network with a learning rate of \verb|0.0001|, $\beta_1=$\verb|0|, and $\beta_2=$\verb|0.9|. The synthetic images are sampled from the trained generator with $128-$dimensional noise vector inputs ($\vec{z}$) that are, in turn, sampled from a Gaussian standard normal  distribution with zero mean and unit variance ($\mathcal{N}(\vec{0}, \mathbbm{1})$) \daggerfootnote{Network architecture and the hyperparameters of the model were chosen out of  those evaluated by Gulrajani \textit{et al.}  on the LSUN bedrooms dataset in \cite{wgan2}. }.

\begin{figure}[tp]
	\centering
	\subfloat[The negative critic loss of the model converging towards a minimum as the network trains. The images used for validation are from a different set of executables not included in the training set.]{
		\makebox[\textwidth][c]{\includegraphics[scale=0.5]{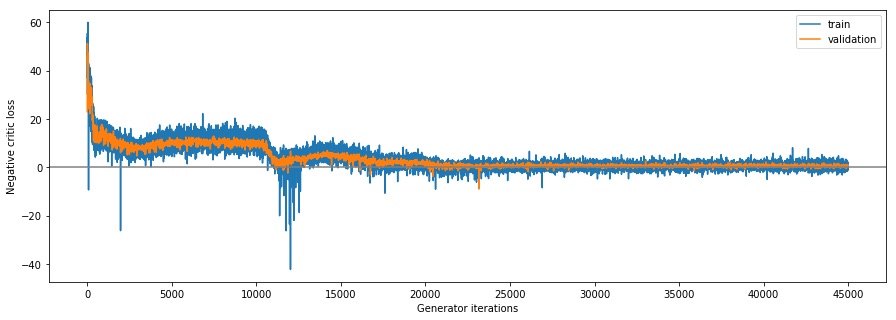}}
		\label{lossplot}
	} \newline
	\centering
	\subfloat[Generated sample of synthetic images from the trained generator after 45,000 generator iterations. (Images shown are scaled.)]{
		\makebox[\textwidth][c]{\includegraphics[scale=0.4]{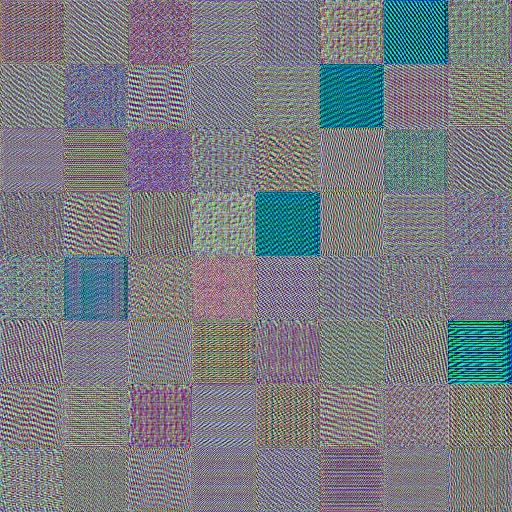}}
		\label{gtbadsynthetic}
	}
	\caption{Training  metrics, and sample generated images of the WGAN-GP model trained on the malicious image dataset. 
	} 
	\label{fig:rm}
\end{figure}

The model trained for about a day across $45,000$ generator iterations with final  negative critic losses (averaged across mini-batches of an epoch) of \verb|0.4262| on the training set, and \verb|1.08365| on the validation set.
Figure \ref{lossplot} describes the training cost (across each iteration) and the validation cost (for every 10 iterations) of the critic on a minibatch of $64$. This shows a stable training run without any overfitting. The Wasserstein distance estimate converges close to zero consistently across the training and the validation. Figure \ref{gtbadsynthetic} shows a sample of $64$ synthetic images generated by the trained generator network $G$.

\subsection{Applications}
\paragraph{Proactive Protection against Zero-Day Malwares.} The synthetic images sampled from the learned distribution, interpolating or extrapolating the known dataset, are representative of new malware behaviors \textit{authored} by the generator. Such generated samples may be  augmented with the original training datasets and can be used for modeling  behavior-based malware classifiers, for protection against zero-day malicious behaviors that are captured by the trained generator's distribution. Malware classifiers that can take such advantage of the synthetic images, needn't necessarily be based only on the image representation. As the  images may be decoded back to the ngram sequence and the corresponding term-frequencies that it visually represents, the synthetic  images  generated, therefore, can further be used by other behavioral event-based rules, signatures and ML applications.

\begin{figure}[tp]
	\centering
\makebox[\textwidth][c]{\includegraphics[width=1.4\textwidth]{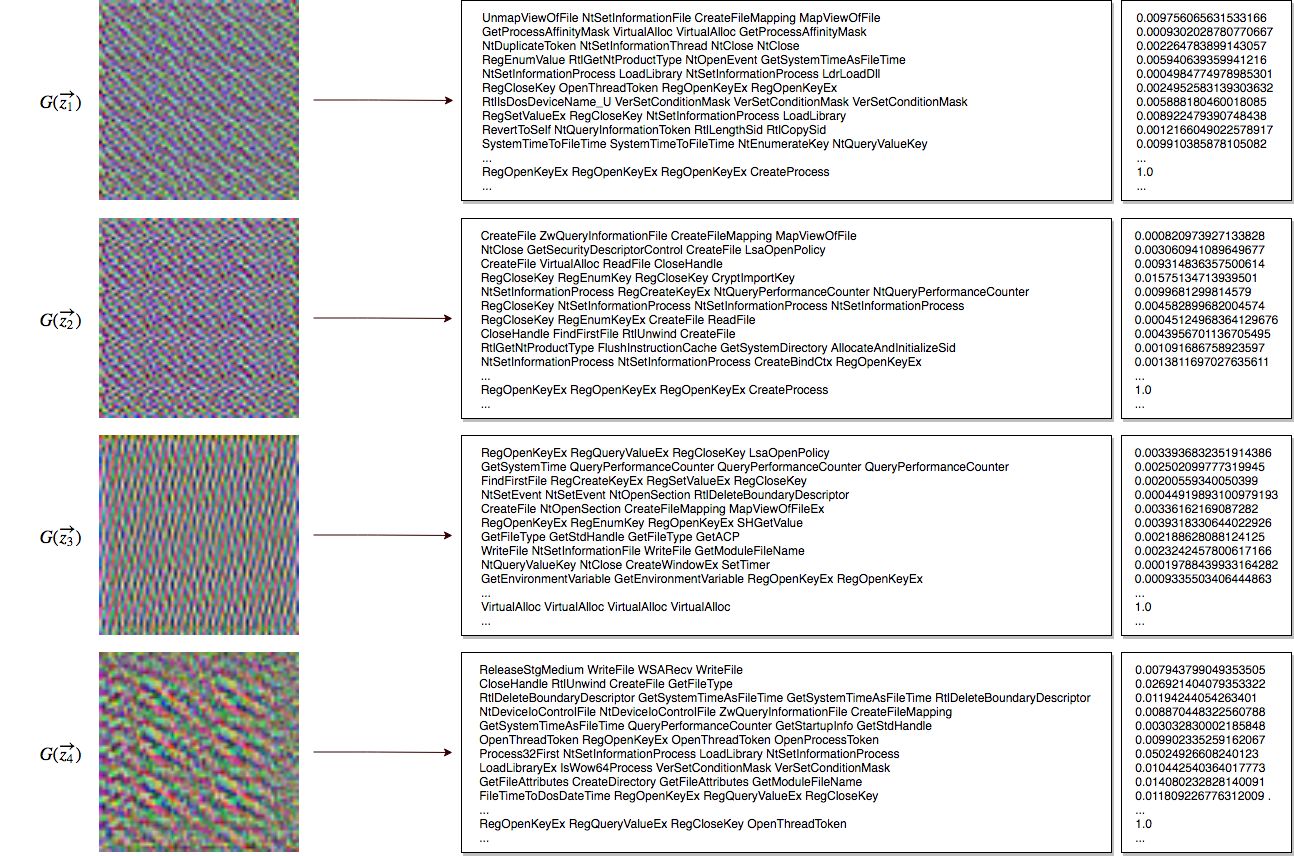}}%
		\caption{Interpreting the synthetic malware by decoding the generated image representation into a sequence of the first-occurrence of the ngram words along with their corresponding relative (sub-linear) term-frequencies with respect to the highest frequent ngram per channel whose tf is normalized to 1. For the demonstration above, $n=4$ (Red channel) is chosen. A similar decomposition may be obtained for 1,2-grams or 3-grams corresponding to the Blue or the Green channels, respectively, as discussed in Section \ref{sec:decoding}. (Images shown are scaled.)}
	\label{syndecode}
\end{figure}

\paragraph{Interpreting the Synthetic Samples.} The synthetic images sampled from the learned distribution may be interpreted for further analysis by decoding them back to the corresponding ngram sequence and the relative term-frequencies, as demonstrated in the Figure \ref{syndecode}. Ability to have such a decomposition offers the advantage of analyzing malicious behaviors that have not yet been observed in-field but are a part of the current threat distribution or \emph{trend} captured by the generator. These decoded sequences may further be used to define behavior-based string-signatures, or augment the training sets of general behavior-based malware classifiers, as discussed above.

\paragraph{Testing and Tuning Malware Classifiers.} Behavior-based malware classifiers may be tuned or tested against the trained generator that takes the role of a malware emulator in synthesizing new malicious behavior, similar to the advantage that the malware authors have been having in being able to test their malicious payload against  security softwares at their disposal.

\paragraph{Generating Malware Hybrids with Vector Arithmetic.} Radford \textit{et al.} \cite{chintala} first showed that the input space of the noise vectors modeled the semantic information of the final generated images in subtle ways that enabled the manipulation of the visual artifacts of the generated image with simple vector arithmetic of the inputs. Specifically, they showed, that manipulating the input noise vectors  corresponding to the output images of \emph{men with glasses}, by intuitive vector arithmetic, gave resultant vectors that sampled  images consisting of \emph{women with glasses}, when fed through the generator $G$ trained on a dataset of human faces. They showed that the image output of $G(\vec{z}_{\text{man,glasses}}-\vec{z}_{\text{man,no glasses}}+\vec{z}_{\text{woman,no glasses}})$ resembled that of a \emph{woman with glasses}. But, a similar arithmetic in the pixel-space, given by $G(\vec{z}_{\text{man,glasses}})-G(\vec{z}_{\text{man,no glasses}})+G(\vec{z}_{\text{woman,no glasses}})$, produced blurry images with all the semantic features of a face washed out. This suggests that the input noise space models the structure of the underlying data in a way, that when combined under an arithmetic operation, samples an image with a combination of specific semantic artifacts (such as the smile, complexion, wearing spectacles etc) while preserving the overall high-level semantics. That is, the resultant generated images from input vectors created through such arithmetic operations, still \textit{looked} like a face. 

\begin{figure}[tp]
	\centering
\makebox[\textwidth][c]{\includegraphics[width=1.5\textwidth]{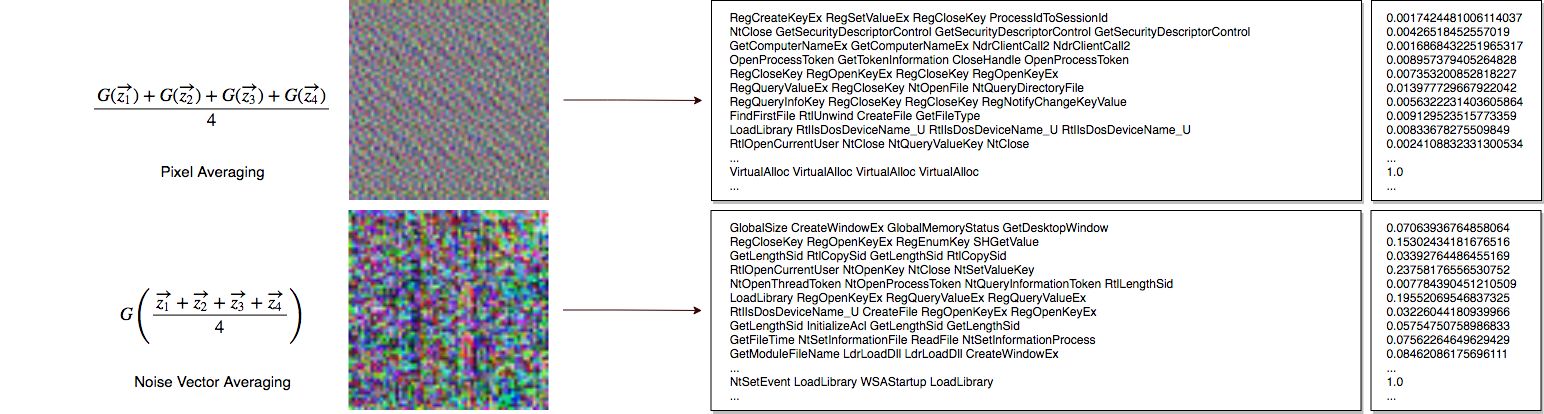}}%
		\caption{Comparing the averages of the malware images in the pixel-domain and the input noise vector space. We propose a novel method of generating hybrid malwares from a set of generated malwares based on  noise vector arithmetic on visual concepts, first demonstrated by Radford \textit{et al.} \cite{chintala} on a dataset of human faces. (Images shown are scaled.)}
	\label{vectorarith}
\end{figure}

This further suggests that such arithmetic in the noise vector space of  malware image representations, may be exploited to combine behavioral features across malwares, such as, for instance, the event of adding an entry to the registry, frequent disk I/O calls infecting good files, using power shell to connect remotely, etc., such that the resultant represents a hybrid malware that is coherently combined to still retain the \textit{malicious} semantics. Figure \ref{vectorarith} demonstrates an example of a simple arithmetic of combining four malwares (those in Figure \ref{syndecode}) in the input noise vector space and the pixel space, highlighting the differences in terms of the decoded  sequences of 4-grams with the corresponding relative term-frequencies.

\section{Conclusions and Future Work}

In this work, we demonstrated a novel way of  representing  dynamic file behavior of an executable in the image-space for effective visualization, and with applications of easy manual/automatic software and malware categorization. The raw pixels of the image representation were shown to encode significant information  to efficiently classify malwares from clean files  using boosted-trees. Additionally, the image representation was shown to be compatible across deep learning architectures (generative or discriminative) involving the convolution filter maps. Finally, we described the applications of GANs in modeling the distribution of malicious behaviors, and in creating new and hybrid malware image representations, that could be decoded to the individual API call information and the first-invocation sequence, for proactive protection.

Future works may include research on incorporating the arguments and the DLL information of the API calls into the images, comparing the results presented in this paper, for possible efficacy improvements. Another possible research direction includes modeling the malware behaviors  directly over the API call sequences using Sequence GANs \cite{seqgans}. Sequence GANs are currently at a nascent state in terms of training stability and sequence generation quality. One may also experiment with the other families of GANs such as the conditional GANs \cite{cgans} that exploit the class labels (malicious or clean) in modeling the class-conditional  distributions of the input data. 

Also, the vector arithmetic heuristic proposed over the image representations of the malware behaviors, for combining malwares, is only easy to formulate, given the vector components capture different recognizable unique strains of the input distribution. That is, each component models a unique  semantic aspect of the malicious behavior. The InfoGAN \cite{infogans} architecture helps to achieve the above by creating a disentangled representation of the semantic aspects, using an augmented set of inputs (called latent code) which, unlike the WGAN-GP model, can be used to deterministically tweak the resultant malware hybrid.

We hope our work  inspires further research into bridging the gap in security with the state-of-the-art deep learning technologies for an end-to-end intelligent proactive protection against unseen and unknown threats.

\end{document}